%% file: main.tex
\documentclass{article} 
\usepackage{latex_style,times}
\input{math_commands.tex}

\usepackage{hyperref}
\usepackage{url}
\usepackage{graphicx}
\usepackage{float}
\usepackage{subcaption} 
\usepackage{cleveref}
\usepackage{ulem}
\usepackage{color}

\title{The Persian Rug: solving toy models of superposition using large-scale symmetries}

\author{Aditya Cowsik\thanks{equal contribution}\\
Physics Department\\
Stanford University\\
{\small \texttt{acowsik@stanford.edu}} \\
\And
Kfir Dolev$^*$  \\
Physics Department\\
Stanford University\\
{\small \texttt{dolev@stanford.edu}}
\And
Alex Infanger$^*$  \\
Independent \\
{\small \texttt{alexdinfanger@gmail.com}}
}

\renewcommand{\vec}[1]{{\mathbf{#1}}}
\newcommand{\relu}{\operatorname{ReLU}}

\newcommand{\ratio}{%
  \mathchoice
    {\frac{n_d}{n_s}}
    {n_d/n_s}
    {n_d/n_s}
    {n_d/n_s}
}
\newcommand{\pf}{p}
\newcommand{\noise}{\nu}
\newcommand{\varvar}{\Delta \operatorname{var}(\noise)}
\newcommand{\Wscale}{a}
\newcommand{\bias}{b}

\newcommand{\gaussmeas}{\frac{e^{-\frac{(\noise+|\mu|)^2}{2\sigma^2}}}{\sqrt{2\pi \sigma^2}}}
\newcommand{\gaussmeasrescaled}{\frac{e^{-\frac{(\noise
+\frac{|\mu|}{\sigma})^2}{2}}}{\sqrt{2\pi}}}
\newcommand{\fullint}{\int_{-\infty}^{\infty}}

\iclrfinalcopy 
\begin{document}

\maketitle

\begin{abstract}

We present a complete mechanistic description of the algorithm learned by a minimal non-linear sparse data autoencoder in the limit of large input dimension. The model, originally presented in \cite{elhage2022superposition}, compresses sparse data vectors through a linear layer and decompresses using another linear layer followed by a ReLU activation. We notice that when the data is permutation symmetric (no input feature is privileged) large models reliably learn an algorithm that is sensitive to individual weights only through their large-scale statistics. For these models, the loss function becomes analytically tractable. Using this understanding, we give the explicit scalings of the loss at high sparsity, and show that the model is near-optimal among recently proposed architectures. In particular, changing or adding to the activation function any elementwise or filtering operation can at best improve the model's performance by a constant factor. Finally, we forward-engineer a model with the requisite symmetries and show that its loss precisely matches that of the trained models. Unlike the trained model weights, the low randomness in the artificial weights results in miraculous fractal structures resembling a Persian rug, to which the algorithm is oblivious. Our work contributes to neural network interpretability by introducing techniques for understanding the structure of autoencoders. Code to reproduce our results can be found at \textcolor{blue}{\href{https://github.com/KfirD/PersianRug}{https://github.com/KfirD/PersianRug}}.
\end{abstract}

\begin{figure}[H]
    \centering
\includegraphics[width=0.4\linewidth]{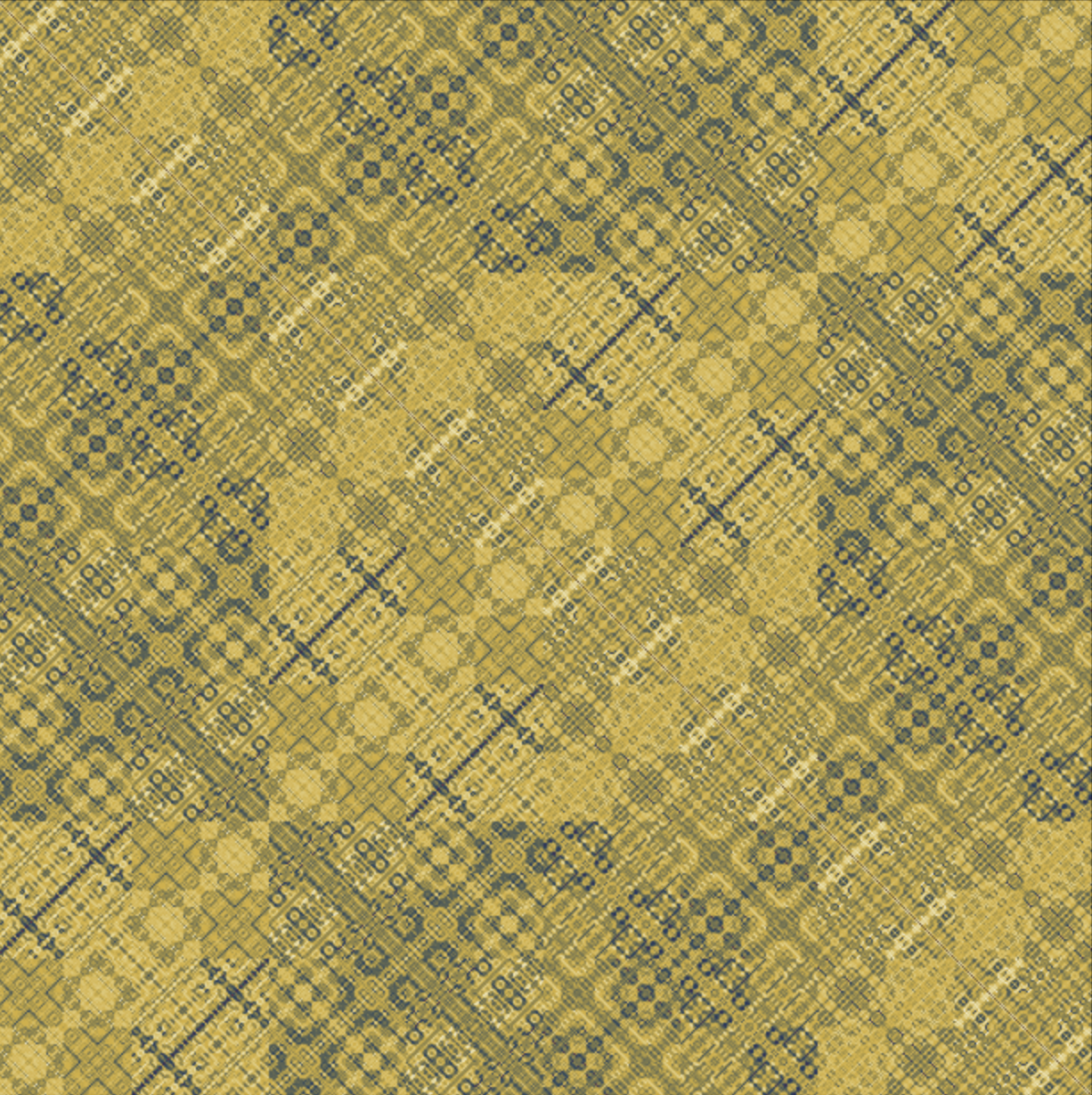}
    \caption{The Persian rug, an artificial set of weights matching trained model performance.}
    \label{fig:persian-rug}
\end{figure}

\section{Introduction}

Large language model capabilities and applications have recently proliferated. As these systems advance and are given more control over basic societal functions, it becomes imperative to ensure their reliability with absolute certainty. Mechanistic interpretability aims to achieve this by obtaining a concrete weight-level understanding of the algorithms learned and employed by these models. A major impediment to this program has been the difficulty of interpreting intermediate activations. This is due to the phenomena of superposition, in which a model takes advantage of sparsity in the input data to reuse the same neurons for multiple distinct features, obscuring their function. Finding a systematic method to undo superposition and extract the fundamental features encoded by the network is a large and ongoing area of research \cite{bricken2023monosemanticity, cunningham2023sparse, gao2024scaling, engels2024languagemodelfeatureslinear}.

Currently, the most popular method of dealing with superposition is dictionary learning with sparse autoencoders. In this method, the smaller space of neuron activations at a layer of interest is mapped to a larger feature space. The map is trained to encourage sparsity and often consists of an affine + ReLU network. This method has been applied to large language models revealing many strikingly interpretable features (e.g. corresponding to specific bugs in code, the golden gate bridge, and sycophancy), even allowing for a causal understanding of the model's reasoning in certain scenarios \cite{marks2024sparsefeaturecircuitsdiscovering}.

The sparse decoding ability of the affine + ReLU map was recently studied in the foundational work \cite{elhage2022superposition}, which introduced and studied a toy model of superposition. The model consisted of a compressing linear layer modeling the superposition\footnote{This is because, if good enough recovery is possible for most features, the pigeonhole principle tells us that at least some of the smaller space activations must encode information about multiple input features.} followed by a decompressing affine + ReLU layer, trained together to auto-encode sparse data. They showed that the network performs superposition by encoding individual feature vectors into nearly orthogonal vectors in the smaller space. The affine layer alone is unable to decode sparse input vectors sufficiently well to make use of superposition, but the addition of the ReLU makes it possible by screening out negative interference. 

While \cite{elhage2022superposition} provides valuable empirical and theoretical insights into superposition, it does not obtain a strong enough description of the model algorithm to quantitatively characterize the algorithm's performance. Given the extensive use of the affine + ReLU map for decoding sparse data in practice, it is important to obtain a complete analytic understanding of the model behavior over a large parameter regime. As we will see, this will inform the design of better sparse autoencoder architectures.

In this work we obtain such an understanding by considering a particularly tractable regime of the \cite{elhage2022superposition} model: permutation symmetric data (no input feature is privileged in any way), and the thermodynamic limit (a large number of input features), while maintaining the full range of sparsity and compression ratio values. In this regime, the learned model weights are permutation symmetric on large scales, which sufficiently simplifies the form of the loss function to the point where it is analytically tractable, leaving only a small number of free parameters. We then forwards-engineer an artificial set of weights satisfying these symmetries and optimizing the remaining parameters, which achieves the same loss as a corresponding trained model, implying that trained models also implement the optimal permutation symmetric algorithm. The artificial set of weights resembles a Persian rug \cref{fig:persian-rug}, whose structure is a relic of the minimal randomness used in the construction, illustrating that the algorithm relies entirely on large-scale statistics that are insensitive to this structure. Finally, we derive the exact power-law scaling of the loss in the high-sparsity regime.

We expect our work to impact the field of neural network interpretability in multiple ways. First, our work provides a basic theoretical framework that we believe can be extended to other regimes of interest, such as structured correlations in input features, which may help predict scaling laws in the loss based on the data's correlations. Second, our work rules out a large class of performance improvement proposals for sparse autoencoders. Finally, our work provides an explicit example of a learned algorithm that is insensitive to microscopic structure in weights, which may be useful for knowing when not to analyze individual weights. 

The paper is structured as follows. In  \cref{sec the-model} we review the model and explain our training procedure. In \cref{sec: empirical-observations} we show empirically that large models display a ``statistical" permutation symmetry. In \cref{sec: extracting-algorithm} we extract the algorithm by plugging the symmetry back into the loss, introduce the Persian rug model which optimizes the remaining parameters, show that large trained models achieve the same loss, and derive the loss behavior in the high sparsity limit. In \cref{sec: related-works} we discuss related works, and conclude with \cref{sec: conclusion}.  




\section{The Model}
\label{sec the-model}
We consider the following model for our non-linear autoencoder with matrix parameters $W_{\text{in}} \in \mathbb{R}^{n_d \times n_s}, W_{\text{out}} \in \mathbb{R}^{n_s \times n_d}, \vec{b} \in \mathbb{R}^{n_s}$,
\begin{align}
f_{\text{nonlinear}}(\vec{x}) &= \relu(W_{\text{out}}W_{\text{in}}\vec{x} + \vec{b}).\label{eq::toy_model}
\end{align}

$W_{\text{in}}$ is an encoding matrix which converts a sparse activation vector $\vec{x}$ to a dense vector, while $W_{\text{out}}$ perform the linear step of decoding. We also consider a simple model for the sparse data on which this autoencoder operates. Each vector $\vec{x}$ is drawn i.i.d. during training, and each component is drawn i.i.d. with $x_i = c_i u_i$, where $c_i \sim \text{Bernoulli}(p)$ and $u_i \sim \text{Uniform}[0, 1]$ are independent variables. This ensures that $\vec{x}$ is sparse with typically only $p n_s$ features turned on. 

We train our toy models to minimize the expected $L_2$ reconstruction loss,
\begin{align}
L(\vec{x}; W_{\text{out}}, W_{\text{in}} , \vec{b}) = E ||\vec{x}-f_{\text{nonlinear}}(\vec{x})||_2^2.\label{eq::basic_loss}
\end{align} 
It is known that for the linear model (\cref{eq::toy_model} without the ReLU), the optimal solution is closely related to principle component analysis (see, for example, \cite{plaut2018principalsubspacesprincipalcomponents} and p. 563 or \cite{bishop2006pattern}). In particular, the reconstruction loss decreases linearly in the hidden dimension $n_d$ when all features are i.i.d. On the other hand, the model \cref{eq::toy_model} will have a much quicker reduction in loss, as will be described in \cref{sec::loss-drop}.

For all models presented we train them with a batch size of $1024$ and the Adam optimizer to completion. That is training continues as long as the average loss over the past 100 batches is lower than the average loss over the 100 batches prior to that one. Our goal with training is to ensure that we have found an optimal model in the large-data limit to analyze the structure of the model itself.

\section{Empirical observations}
\label{sec: empirical-observations}

In this section, we present empirical observations of the trained models. We start by presenting a remarkable phenomenon this model exhibits in the high-sparsity regime: a dramatic decrease in loss as a function of the compression ratio. We then turn to a mechanistic interpretation of the weights which gives empirical evidence for the phenomena needed to understand the algorithm the model learns. These are manifestations of a partially preserved permutation symmetry of the sparse degrees of freedom. 

\subsection{Fast loss drop}\label{sec::loss-drop}
\begin{figure}[H]
    \centering

    \includegraphics[width=0.48\linewidth]    {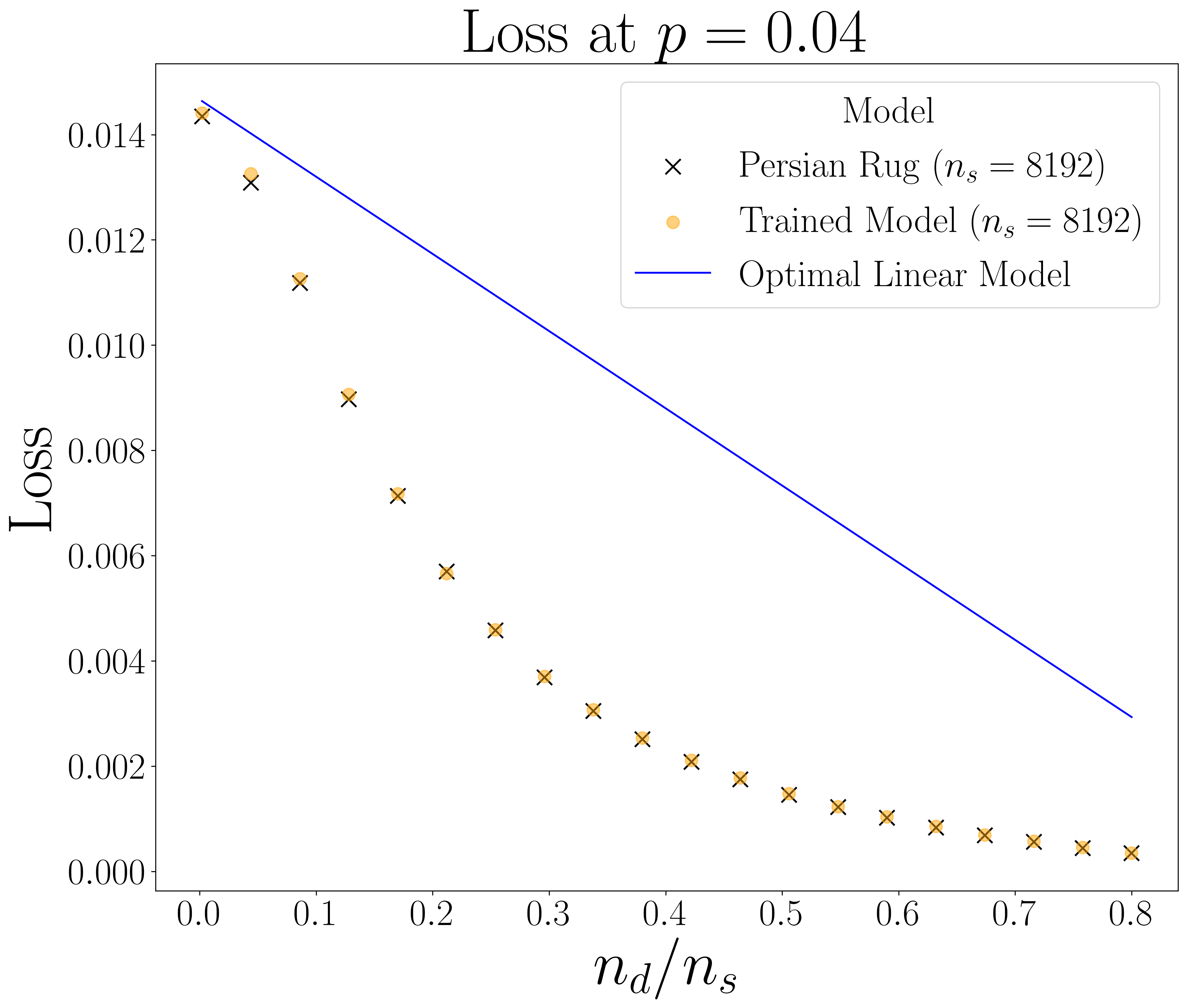}
    \caption{Loss curves of trained models, Persian rug models, and optimal linear models as a function of the compression ratio.}
    \label{fig:loss-curve}
\end{figure}

To gauge the performance of the model, we plot the loss (\cref{eq::basic_loss}) as a function of the compression ratio $\ratio$. 
%
 In \cref{fig:loss-curve} we plot the performance of the linear versus non-linear models for representative parameters. It is clear that the non-linear model outperforms the linear model up until near the $\ratio\approx 1$ regime due to an immediate drop in the loss.
 The slope and duration of this initial fall is controlled by $\pf$. In particular, in the high-sparsity regime ($\pf$ close to zero), the loss drops to zero entirely near the $\ratio\approx 0$ regime. What is going on here? To explain this behavior, we analyze the algorithm the model encodes.

\subsection{Statistical permutation symmetry}
\label{sec:stat_perm_symmetry}
\begin{figure}[H]
    \centering

    \includegraphics[width=0.55\linewidth]{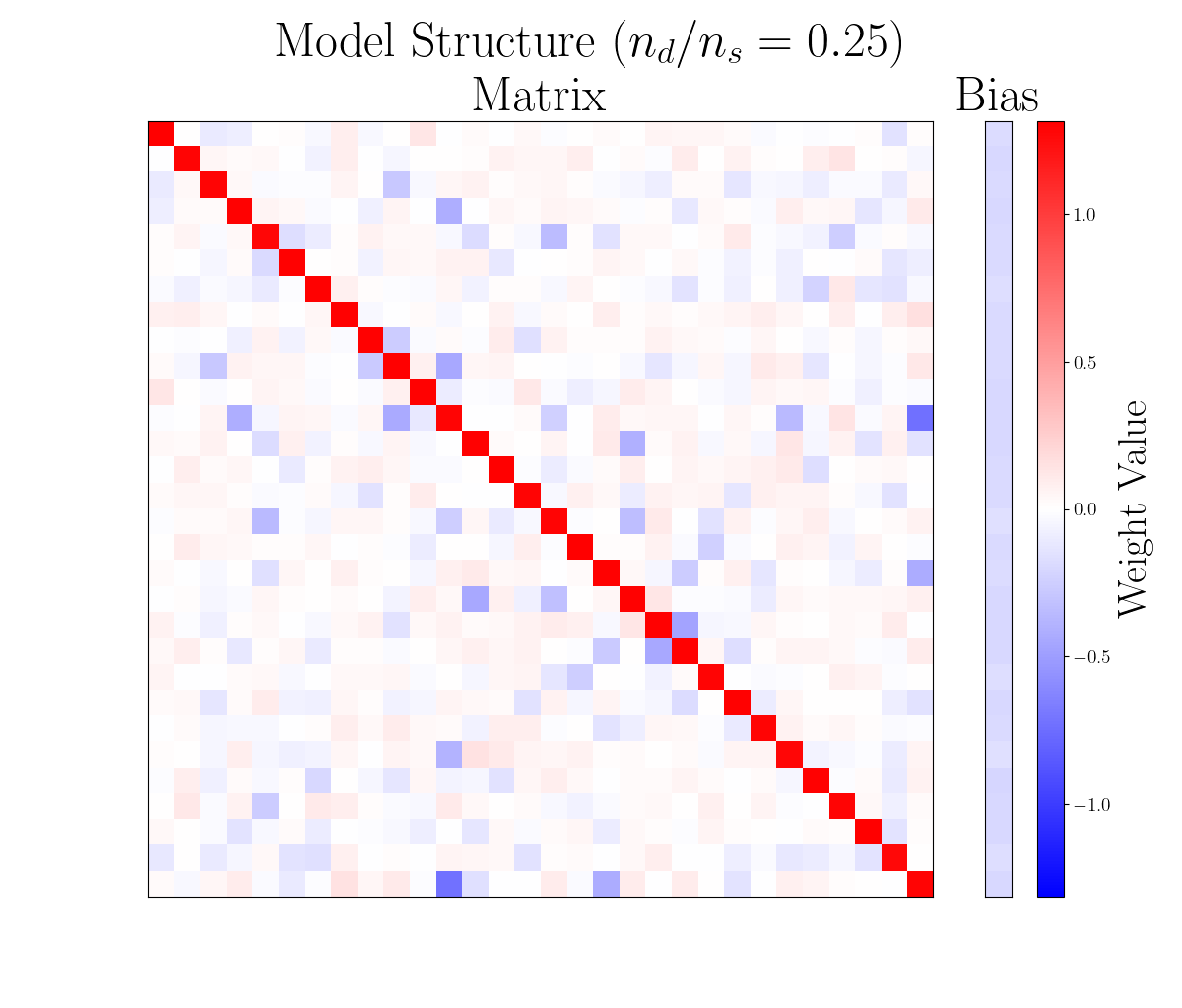}
    \caption{Plot of the first $30\times 30$ $W$ elements and the corresponding bias ($\vec{b}$) components, at $\pf = 4.5\%$ and ratio $n_s = 512$. The diagonal components are all at similar values of $1.29 \pm .01$ (one standard deviation) while the off-diagonal components are approximately mean-zero, appearing like noise. The bias elements are all negative around $-.18 \pm .01$. This statistical uniformity is a permutation symmetry across the sparse features.}
    \label{fig:W-matrix}
\end{figure}

\begin{figure}[H]
    \includegraphics[width=\textwidth]{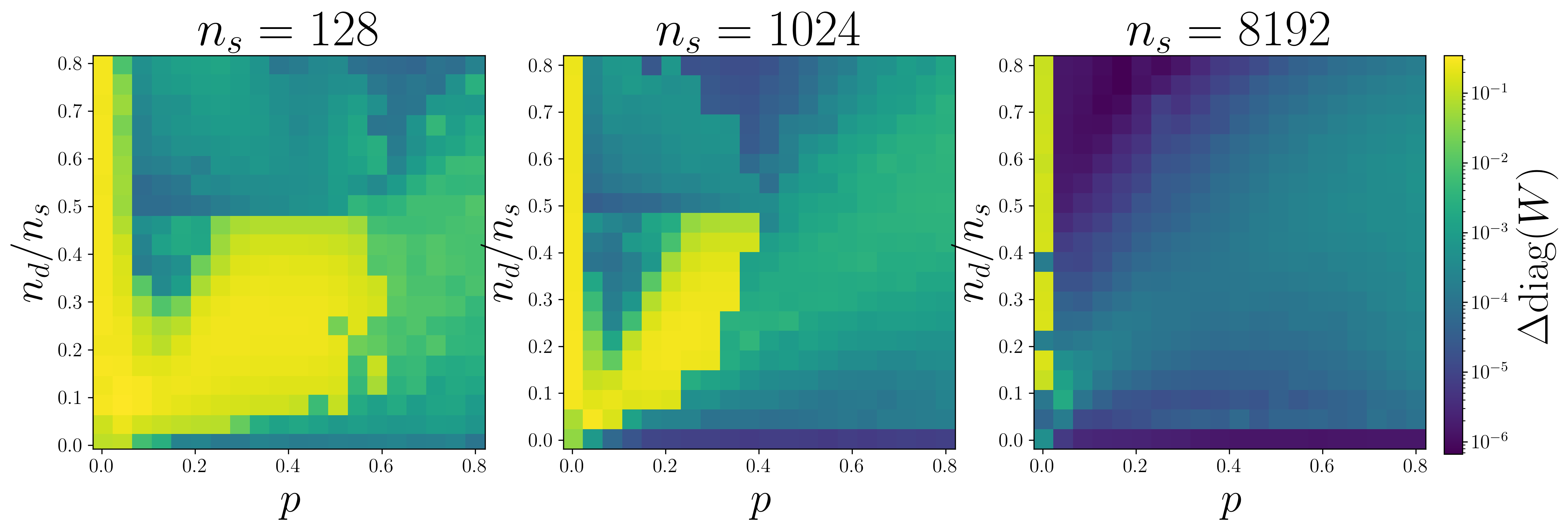}
    \caption{Permutation symmetry of diagonal values. We plot the mean-square fluctuation of the diagonal values corresponding to each model. Models are trained as a function of $\pf$ and $\ratio$. The emergence of symmetry as $n_s$ grows (at all locations in the diagram) is a crucial element of the algorithm implemented by the autoencoders.}
    \label{fig:empirical-diagonal-concentration}
\end{figure}

\begin{figure}[H]
    \centering
    \includegraphics[width=\textwidth]{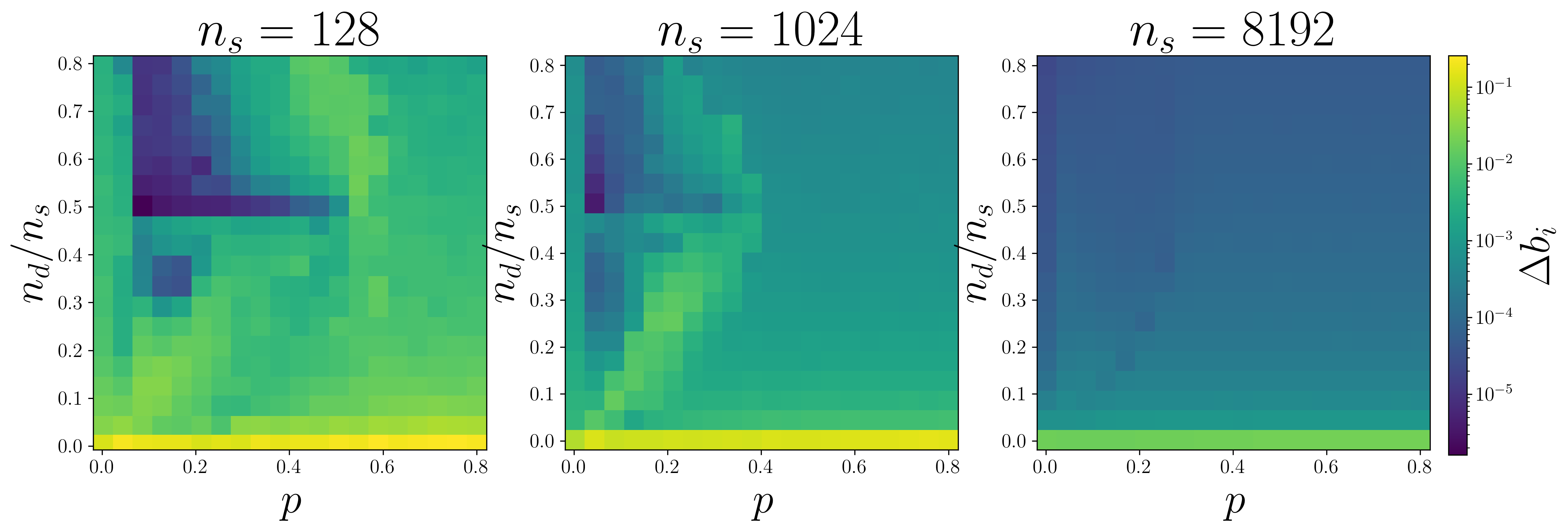}
    \caption{Permutation symmetry of bias values. We plot the mean-square fluctuation of values in the bias vectors corresponding to each model, which are trained as a function of $\pf$ and $\ratio$. As $n_s$ increases the fluctuation over bias elements generally decreases in all trained models.}
    \label{fig:empirical-bias-concentration}
\end{figure}

\begin{figure}[H]
    \centering
    \includegraphics[width=\textwidth]{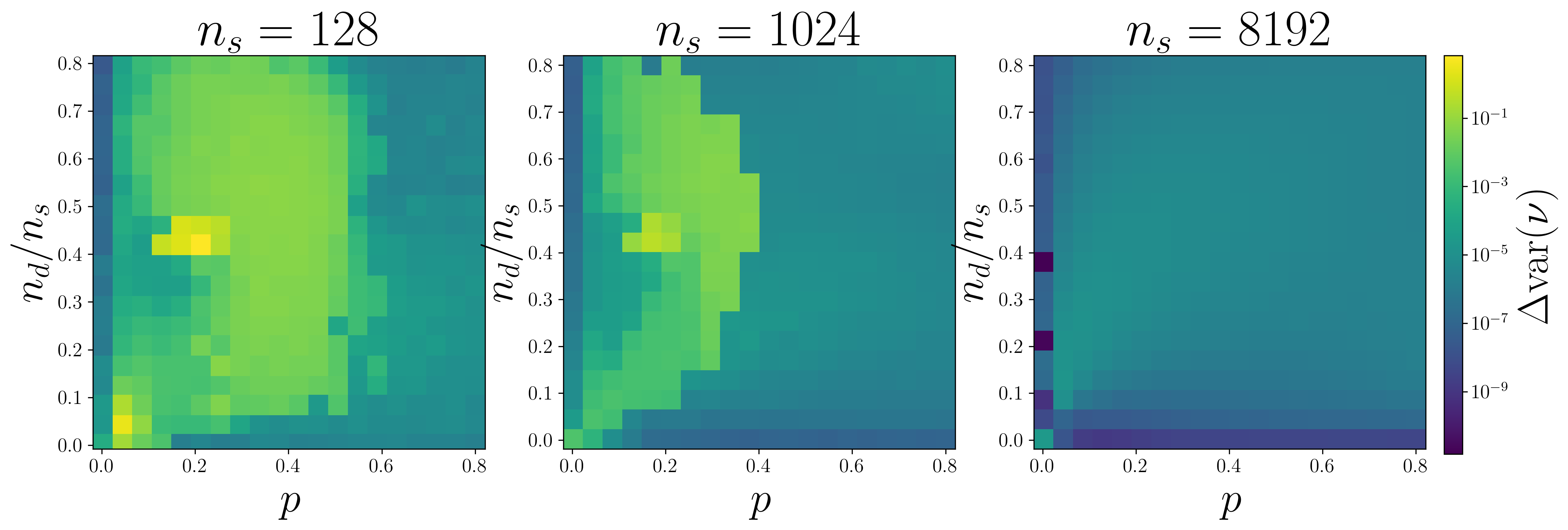}
    \caption{Permutation symmetry of off-diagonal statistics. The symmetry breaking parameter $\varvar$ is given by the variance across all rows of the squared sum of the off diagonal elements in each row, up to a constant. Once $n_s$ reaches $8192$ all noise variables have nearly identical variances.}
    \label{fig:empirical-off-diagonal-symmetry}
\end{figure}

\begin{figure}[H]
    \centering
    \includegraphics[width=\textwidth]{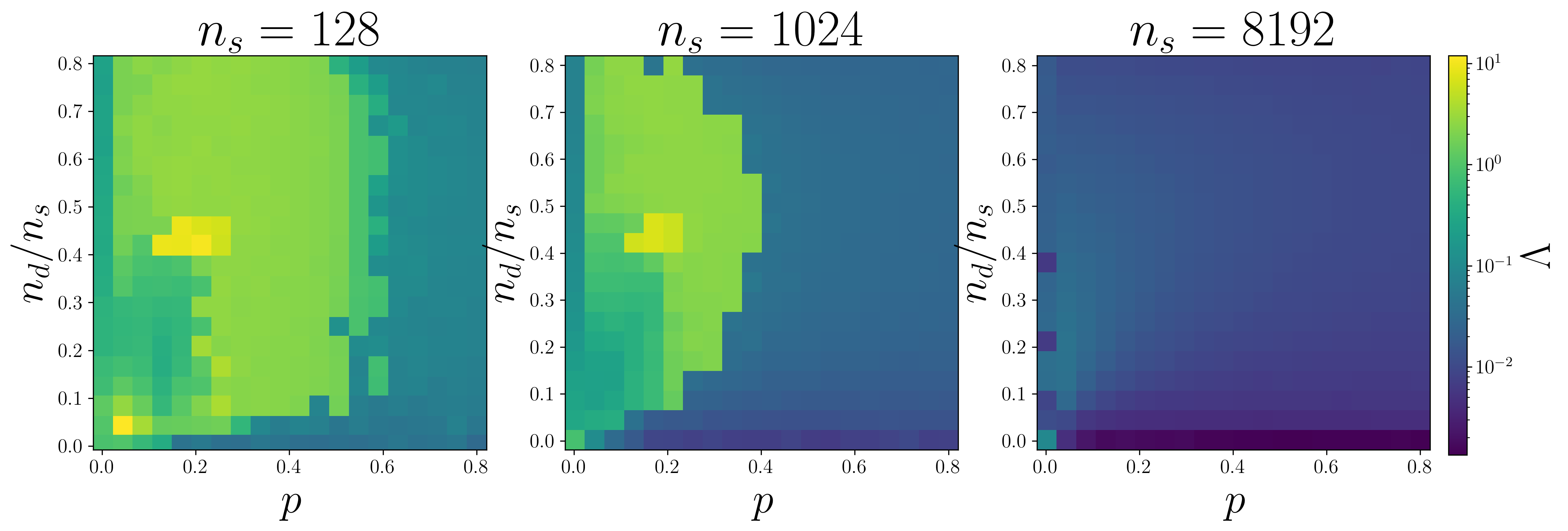}
    \caption{Interference Gaussianity in the large $n_s$ limit. We show that the Lyapunov condition is better satisfied as $n_s$ grows larger, by plotting the worst-case Lyapunov condition (see \cref{eq:lyapunov_condition}) over all rows of $W$. Models are trained at different $\pf$ and $\ratio$ values. We see that for small $n_s$ only models trained at large $\pf$ have nearly-uniform off-diagonal entries whereas all models approach uniformity at large $n_s$.}
    \label{fig:empirical-gaussianity}
\end{figure}

Rather than looking individually at the weights, it is helpful to look at the matrix $W=W_{\text{out}}W_{\text{in}}$, shown in \cref{fig:W-matrix}. As will be made precise in \cref{sec: extracting-algorithm}, we are interested in the structure of each row since it is responsible for the corresponding feature output. Though the matrix is not permutation symmetric in detail, its statistics are sufficiently so as far as the algorithm is concerned. In particular, in the $\ratio\rightarrow\infty$ limit, the matrix is statistically permutation invariant in the following sense: 
\begin{enumerate}
    \item the diagonal elements become the same (\cref{fig:empirical-diagonal-concentration}), 
    \item the bias elements become uniformly negative, which can be seen in the uniformity and slight blue shade in \cref{fig:W-matrix} and is quantified in \cref{fig:empirical-bias-concentration},
    \item the statistics of the off-diagonal terms in each row are sufficiently uniform for interference to become Gaussian (\cref{fig:empirical-gaussianity}), and finally
    \item the mean and variance of each of these becomes the same across rows (\cref{fig:empirical-off-diagonal-symmetry}).
\end{enumerate}

Immediately on looking at the exemplar model from \cref{fig:W-matrix} we see that the diagonals are large and uniform while the off-diagonal components are fluctuating and much smaller. This serves the function of routing each input to the corresponding output, except with some off-diagonal contributions which should be as small as possible. 

Because the diagonal components appear uniform, we quantify this by looking at the fluctuation on a per-model basis. For each model, we compute the empirical mean and variance of the diagonal components and plot the latter in \cref{fig:empirical-diagonal-concentration}. Initially, when $n_s = 128$ is small, we see a region with large fluctuations among the diagonal components, particularly when the $\pf$ and $\ratio$ are both small. We hypothesize this occurs because for small models (and small feature probabilities) the contribution to any output from off-diagonal terms may fluctuate because of the small size of the matrix.

We see similar features among the entries of the bias in \cref{fig:empirical-bias-concentration}. The relatively large initial fluctuations generally decay as $n_s$ grows larger. Similar features are present in the two plots, such as the relatively low fluctuation regions near $(\pf, \ratio) \approx (.1, .5)$ because the bias and the diagonal entries are coupled when a diagonal element in a row is large the corresponding bias must be scaled up as well and vice versa.

For each of these algorithm-revealing features, we give three plots, each with increasing $n_s$, to show that the feature becomes apparent in the thermodynamic limit. Each plot is a heat map with the variable representing the feature plotted in color and spatial dimensions covering the entire parameter space of $\ratio$ and $\pf$.

We similarly see in \cref{fig:empirical-off-diagonal-symmetry} that the off-diagonal terms concentrate as well. To be more precise, we look at the fluctuation of the total norm of the off-diagonal elements in each row in $W$ across rows. The norm is a measure of the amount of signal arriving at a particular output from other inputs. A smaller contribution from other inputs means that each output has a higher signal-to-noise ratio as long as the diagonal term remains constant. 

Given that there are many off-diagonal terms, and those terms are all added together in the process of multiplying $W \vec{x}$, it is natural to ask if they satisfy the Lyapunov condition, which would imply that the contribution to output $i$ from off-diagonal terms $j \neq i$ is Gaussian. In \cref{fig:empirical-gaussianity} we plot the statistic 
\begin{equation}
    \Lambda \equiv \max_{i} \frac{\sum_{j\neq i} |W_{ij}|^3}{\left(\sum_{j\neq i} W_{ij}^2 \right)^{3/2}}
    \label{eq:lyapunov_condition}
\end{equation}
where the sum is over all values of $j \neq i$ for each $i$. The statistic tends towards 0 as $n_s$ grows larger which means that the contribution from the off-diagonal elements tends towards a Gaussian distribution. This is strong numerical evidence that no small set of terms in the off-diagonals dominate.

\subsection{Optimization of residual parameters}
The statistical permutation symmetry places constraints on the possible values of $W$ and $\vec{b}$. The constraint on $\vec{b}$ is straightforward: it is proportional to the all ones vector, i.e. there is a number $\bias$ such that $\vec{b}_i=\bias$ for all $i$. As will be explained in \cref{subsec:trained-models-are-optimal}, the relevant degrees of freedom remaining in $W$ are one number $\Wscale$ equal to the diagonals, and $\sigma$ characterizing the root mean square of the off diagonal rows. The precise values of the off diagonals can be thought of as irrelevant ``microscopic information". Thus there are three relevant degrees of freedom remaining: $b,a$, and $\sigma$.

In \cref{subsec: persian-rug-model} we give a specific set of values for $W_{\text{in}}$ and $W_{\text{out}}$ via the ``Persian rug" matrix, which have the statistical permutation symmetry in the $n_s\rightarrow\infty$ limit while also optimizing $\sigma$. The remaining parameters, $a$ and $b$ can be optimized numerically. In \cref{fig:loss-curve} we compare the loss curve of this artificial model with that of a trained model, and see that they are essentially the same.

\section{Extracting the Algorithm}
\label{sec: extracting-algorithm}
In this section we give a precise explanation of the algorithm the model performs. We start with a qualitative description of why the statistical permutation symmetry gives a good auto-encoding algorithm when the remaining macroscopic degrees of freedom are optimized. We then find an artificial set of symmetric weights with optimized macroscopic parameters. We show that the trained models achieve the same performance as the artificial model, thus showing they are optimal even restricting to statistically symmetric solutions. Finally, we derive an explicit form of the loss scaling and argue that ReLU performs near optimally among element-wise or ``selection" decoders. 


\subsection{Qualitative Description}
\label{subsec:trained-models-are-optimal}
A key simplification is to consider strategies as collections of low-rank affine maps rather than as the collection of weights directly. In other words consider the tuple $(W, \vec{b})$ where $W = W_{\text{out}}W_{\text{in}}$ to define the strategy. We must restrict to $W$ with rank no more than $n_d$ because it is the product of two low-rank matrices. Given any such $W$ we may also find $W_{\text{in}}$ and $W_{\text{out}}$ of the appropriate shape (e.g. by finding the SVD), so the two representations are equivalent. 

Let us now take a close look at the reconstruction error. Let us consider output $i$, which is given by
\begin{equation}
    (f_\text{nonlinear}(\vec{x}))_i 
    =\relu(W_{ii}x_i + \sum_{j=1, j\neq i}^{n_s} W_{ij} x_j + \vec{b}_i) 
    = \relu\left(a(x_i + \noise_i) + \vec{b}_i\right)
\end{equation}
where we used that all diagonal elements of $W$ are equal to $a$ and defining $\noise_i=\Wscale^{-1}\sum_{j=1, j\neq i}^{n_s} W_{ij} x_j$. At this stage we can already see that the individual off-diagonal elements are only important insofar as they characterize the Gaussian random variable $\noise_i$. More specifically, because $\noise_i$ is completely characterized by its mean and variance, only the sum and sum of squares of the off-diagonal components matter. Because we have a bias term the mean of $\noise_i$ may be absorbed into it, so from now we assume that $\nu_i$ are all zero-mean. This simplification is why permutation symmetry emerges when $n_s$ is large, and allows us to consider only the macroscopic variables introduced when we discuss the loss.

The reconstruction error is (dropping the $i$ subscript, and substituting $\vec{b}_i = b$)
\begin{align}
    L = \mathbb{E}_{x,\noise}[(x-\relu\left(a(x + \noise) + b\right))^2].
    \label{eq:expected_loss}
\end{align}
We can further decompose this by taking the expectation value over whether $x$ is on or off, so
\begin{align*}
    L = (1-p)L_\text{off} + pL_\text{on}
\end{align*}
where 
\begin{align*}
    L_\text{off} &= \mathbb{E}_{\noise}[\left(\relu\left(a\noise + b\right)\right)^2], \text{ and}\\
    L_\text{on} &= \mathbb{E}_{u,\noise}[(u-\relu\left(a(u+\noise) + b\right))^2]
\end{align*}
with ${u\sim\text{Uniform}[0,1]}$. Let us explore the regimes of macroscopic parameters $a,b,\sigma$ when either or both of these loss terms are low.

The impact of all the non-diagonal terms has been summed up in the ``noise'' $\noise$. Though it is a deterministic function of $\vec{x}$, output $i$ has no way to remove $\nu$ from $u + \nu$. The best it can do is to estimate $u$ from $u + \nu$ because it doesn't have any other information. This exemplifies a key principle -- by restricting the computational capacity of our decoder, deterministic, but complicated correlations act like noise.

The main advantage of the nonlinear autoencoder is that the dominant contribution to the loss, $L_\text{off}$, can be immediately screened away by making $a\noise + b$ either small or negative, allowing the network to focus on encoding active signals. This immediate screening is always possible either by choosing $b+\mu$ large and negative or $a$ and $\mu+b$ small. However, these strategies come at a cost because the output value is distorted from $u$ to $au+b$. It is thus preferable instead for $\noise$ and $b$ to be as small as possible, which occurs when $\sigma$ is as small as possible. As we will see, $\sigma$ is the only parameter we are not free to choose in the large $n_s$ limit, and whose value will be bounded as a function of $\ratio$ and $\pf$. Since $L_\text{off}$ is the dominant contribution to the loss, therefore, it will thus be necessary to damp the signal by setting $\Wscale$ small and/or $b$ large and negative in regimes where $\sigma$ is uncontrollably large.







Given that we see a statistical permutation symmetry in trained models let's consider symmetric strategies so that $W_{ii} = \Wscale$ and $\vec{b}_i=b$ for all $i = 1, \dots, n_s$. We will show that optimizing the remaining macroscopic parameters makes $f_\text{nonlinear}$ act close to the identity on sparse inputs. 

\subsection{Optimizing the macroscopic parameters}
\label{subsec: opt-mac-params}

We have seen qualitatively that a statistically symmetric strategy exists in certain regimes of the macroscopic parameters. Two of these parameters, $\Wscale$ and $b$ are unconstrained. Furthermore the loss should be monotonically increasing with $\sigma$ because a larger $\sigma$ implies more noise which hinders reconstruction. Thus we now prove lower bounds on $\sigma$ and construct an artificial set of statistically permutation symmetric weights which achieve this bound. Finally we will compare the reconstruction loss of this strategy with the learned one to justify that those ignored microscopic degrees of freedom were indeed irrelevant.

\label{subsec: optimizing-parameters}
\subsubsection{Optimal $\sigma$}

Assuming the permutation symmetry we discovered earlier in our empirical investigations we will derive a bound on the variance of the output. Additionally for an optimal choice of $\vec{b}$ the average loss is increasing in the variance, because a larger variance corresponds to a smaller signal-to-noise ratio. Taken together these two facts will give a lower bound on the loss. We will then provide an explicit construction which achieves this lower bound and illustrates how the algorithm works.

The lower bound on the variance comes from the fact that $W$ is low-rank with constant diagonals. For now let us ignore the overall scale of $W$, and just rescale so that the diagonals are exactly 1. The bound we are about to prove is very similar to the Welch bound \citep{welch1974lower} who phrased it instead as a bound on the correlations between sets of vectors. We produce an argument for our context, which deals with potentially non-symmetric matrices $W$, the details of which are located in \cref{app:min_variance}.

We show that 
\begin{equation}
\label{eq:sigma_bound}
    \sigma^2 \geq \frac{4 p - 3p^2}{12}\left(\frac{n_s}{n_d}-1\right) 
\end{equation}
with equality only when $W$ is symmetric, maximum rank, with all non-zero eigenvalues equal. This naturally leads to a candidate for the optimal choice of $W$, namely matrices of the form
\begin{equation}
    W \propto O P O^T \text{ and } W_{ii} = 1
\end{equation}
where $O$ is an orthogonal matrix and $P$ is any rank-$n_d$ projection matrix. This kind of matrix saturates the bound because it is symmetric and has all nonzero eigenvalues equal to 1. 

A note on the connections between these matrices and tight frames -- if we take the ``square root'' of $W$ as a $n_s \times n_d$ matrix $\sqrt{W}$ such that $\sqrt{W}\sqrt{W}^\dagger = W$ then the $n_d$ dimensional rows of $\sqrt{W}$ are a tight-frame. This is because $\Tr(W)^2 = n_s^2 = n_d \Tr WW^\dagger$ which is the variational characterization of tight-frames as in Theorem 6.1 from \citet{waldron2018introduction}. 

\subsubsection{Persian rug model}
\label{subsec: persian-rug-model}

We now give an explicit construction of an optimal choice for $W$. The construction is based on the Hadamard matrix of size $n_s = 2^m$ for some integer $m$, defined as
\begin{align*}
    H_{ij} = \operatorname{parity}(\texttt{bin}(i) \oplus \texttt{bin}(j))
\end{align*}
where $\texttt{bin}(i)$ is the $m$ bit binary representation of $i$ and $ \oplus$ is the bit-wise xor operation. We then define a Persian rug matrix as
\begin{align*}
    R = n_d^{-1}\sum_{k \in S} H_{ik}H_{kj}
\end{align*}
where $S\subset \{1,...,n_s\}$ with $|S|=n_d$. In \cref{fig:persian-rug} we plot such a matrix for $n_s = 256$, $n_d=40$, and $S$ chosen randomly. The matrix $R$ has diagonals equal to 1 because each diagonal is the average of $n_d$ terms equal to $(\pm 1)^2$, and a projector because the rows of $H$ are orthogonal, and thus saturates \cref{eq:sigma_bound}.  Furthermore, one can readily check numerically that it satisfies the statistical symmetries. There remain two variables to optimize, $\Wscale$ and $\bias$ (recall $\mu$ can be absorbed into $b$). We do this numerically and compare to a trained model in \cref{fig:loss-curve}. 

\subsection{Loss scaling at high sparsity}
\label{sec: loss_scaling}

Having obtained a simple expression for the loss in terms of constants $a,b$ and two simple random variables $x\sim \text{Uniform}[0,1]$ and $\noise\sim \mathcal{N}(\mu,\sigma)$, as well has having deduced an achievable lower bound for $\sigma$, we are now able to explain why the simple ReLU model performs so well at high sparsity. For ease of notation let us use $r = \ratio$. 

\subsubsection{Initial loss (ratio=0)}
 
Let us first consider the $r \rightarrow0$ limit with all other parameters fixed. Then $\sigma\rightarrow\infty$ because of the bound in \cref{eq:sigma_bound} so the fluctuations in $\noise$ overwhelms the signal term. This means that the optimal $a$ is
\begin{equation}
    a = 
    p\frac{\mathbb{E}_{u,\noise}\left[u \relu(\noise + \bias)\right]}
    {\mathbb{E}_{\noise} \left[ \relu(\noise + \bias)^2 \right]} + O(\sigma^{-1}).
\end{equation} 
The loss then becomes 
\begin{align*}
    L& = (1-p)a^2\mathbb{E}_{\noise}[\relu\left(\noise + \bias \right)^2]
    +
    p\mathbb{E}_{u,\noise}[(u-a\relu\left(\noise + \bias)\right)^2] + O(\sigma^{-1})\\
   & =
   a^2\mathbb{E}_{\noise}[\relu\left(\noise +\bias\right)^2] - 2ap\mathbb{E}_{u,\noise}[u\relu(\noise + \bias)] + p\mathbb{E}_{u}[u^2] + O(\sigma^{-1})
\end{align*}
plugging in $a$ explicitly gives
\begin{align*}
        L = p\mathbb{E}_{u}[u^2] 
        -
        p^2\frac
        {\left(\mathbb{E}_{u}\left[u \right]\right)^2
        \left(\mathbb{E}_{\noise}\left[\relu(\noise + \bias)\right]\right)^2}
        {\mathbb{E}_{\noise} \left[ \relu(\noise + \bias)^2 \right]} + O(\sigma^{-1}).
\end{align*}
Thus we can conclude that
\begin{align*}
    \lim_{r\rightarrow 0 } L = p \mathbb{E}_{u}[u^2]  + O(p^2) = \frac{p}{3}  + O(p^2)
\end{align*}

Thus we see that in the $p\ll1$ regime we have $ L\rightarrow L_0(p) \sim O(p)$ independent of the other parameters. We will now see that increasing $r$ will quickly cause the loss to drop to $O(p^2)$. 

\subsubsection{Loss upper bound}
We now show that the loss drops off quickly in the sense that for  $\frac{r}{p}\gg 1$ we get that $L(p)/p \rightarrow 0$, i.e. $L(p)$ scales super-linearly with $p$. We will consider the regime where $r\ll 1$ holds\footnote{For example $r= p^{1-\epsilon}$ for any $\epsilon\in (0,1)$.}  so that we may take $\sigma^2\sim \frac{p}{r}\ll1$.

To obtain the upper bound we will make educated estimates for values of $a$ and $\mu$ that are near optimal. In particular, in \cref{apx: optimizing-A} we show that the optimal value of $a$ is (after absorbing $b$ into the mean of $\nu$, $\mu$) :

\begin{equation}
    a_{\text{opt}} = \frac{\mathbb{E}_{x, \noise}\left[x \relu(x+\noise)\right]}{\mathbb{E}_{x, \noise} \left[ \relu(x + \noise )^2 \right]}.
\end{equation}
From the form of the loss, we know that $\mu$ must decrease as $p$ decreases for the loss to go down faster than $O(p)$. Thus $\nu$ has both a mean and variance approaching 0, and $a_\text{opt}\rightarrow 1$. Thus we plug in $a=1$ before taking these limits in the expectation of getting a good upper bound. The loss then takes the form
\begin{align*}
    L = (1-p)L_\text{off} + pL_\text{on} 
\end{align*} 
with 
\begin{align*}
    L_\text{off} &= \mathbb{E}\left[\relu\left(\noise\right)^2\right], \text{ and}\\
    L_\text{on} &= \mathbb{E}\left[(u-\relu\left(u+\noise)\right)^2\right]
\end{align*}

\paragraph{Off term:}

The off term can be upper bounded via
\begin{align}\label{eq: relu-chi-2}
    \mathbb{E}\left[\relu(\noise)^2\right] 
    &= \int_0^\infty d\noise \gaussmeas\noise^2 
    = \sigma^2\int_0^\infty d\noise \gaussmeasrescaled\noise^2\\
    &\leq \frac{\sigma^2}{2} e^{-\frac{\mu^2}{2\sigma^2}}
\end{align}
and thus we see we need to set $\frac{\mu}{\sigma}\gg 1$ to get a good bound. In particular, we know empirically that the loss drop happens at increasingly smaller $r$. To ensure this we let $\sigma^2\sim \frac{p}{r}$ scale at some rate slower than $p$. Thus to ensure that the total loss decreases faster than $O(p)$, we need $e^{-\frac{\mu^2}{2\sigma^2}}\sim O(p)$ or in other words 
\begin{align}\label{eq: mu-scaling}
    \mu\sim\sigma\sqrt{\log{\frac{1}{p}}}. 
\end{align}

\paragraph{On term:} We perform a similar, but slightly more involved computation in \cref{app:reconstruction_error} and combine with the off term to obtain 
\begin{align*}
    L < 
    (1-p)O(\sigma^2 e^{-\frac{\mu^2}{2\sigma^2}}) + 
    pO(
    \mu^3 
    + 
    e^{  -\frac{|\mu|^2}{2\sigma^2}    } 
    + 
    \mu e^{  -\frac{|\mu|^2}{2\sigma^2}    } 
    + 
    \sigma^2 
    + 
    \sigma\mu 
    + 
    \mu^2).
\end{align*}
Plugging in the $\mu$ scaling from \cref{eq: mu-scaling} and keeping only the lower order terms gives
\begin{align}
    L < O\left(
     \sigma^2 p\log{\frac{1}{p}} \right) \sim O\left(\frac{p^2}{r}\log{\frac{1}{p}}\right).
\end{align}
\subsubsection{Loss lower bound}
In \cref{apx: lower_bounding_loss} we also derive a lower bound in the high-sparsity limit $ L > O(\frac{p^2}{r})$
in the high sparsity limit up to logarithmic corrections. We show this in fact holds for a more general class of activation functions. In particular, any function which acts element-wise or filters out elements will give an on-loss contribution of the form
\begin{align*}
    \mathbb{E}_{u,\nu}[(u-f\left(u+\noise)\right)^2]
\end{align*}

which has a lower bound due to $\nu$ destroying information about $u$. Thus we can conclude that 
\begin{align*}
    L \sim O\left(\frac{p^2}{r}\right)
\end{align*}

up to logarithmic factors whenever $\frac{p}{r}\ll 1$.


\section{Discussion}
\label{sec: conclusion}

\subsection{Summary}

We have performed an exhaustive analysis of the toy model of superposition for permutation symmetric data in the limit of large number of sparse signals. We empirically showed that the network reliably learns statistically permutation symmetric algorithms, in the sense that the information relevant to the algorithm is encoded in the summary statistics of the elements in each row of the $W$ matrix. 

Plugging these symmetries back into the reconstruction-loss allowed us to reduce its optimization problem over all model parameters to an optimization problem over three scalar variables. Of these parameters, only the noise parameter $\sigma$ was constrained by the hyper-parameters $p,n_s,$ and $n_d$. We showed that $\sigma$ is minimized for a statistically symmetric $W$ that is proportional to a projector.

We then forward-engineered weights optimizing these parameters, giving the Persian rug model, and showed empirically that its loss matched the model loss closely. This shows that the trained models learn the optimal symmetric algorithm. The additional presence of  small scale structure in the Persian rug model highlights the algorithm's independence on such smaller scale details.  

Finally, we considered the analytic form of the loss in the high-sparsity limit, and derived its scalings as a function of sparsity and compression ratio to be $O(p)$ when $\frac{r}{p}\ll 1$ and $\sim O(p^2/r)$ when $\frac{p}{r}\ll 1$, or in other words the loss drops of $O(p)$ in a ratio window scaling faster than $O(p)$. 

\subsection{Relationship to other works}
\label{sec: related-works}


\subsubsection{Mechanistic Interpretability and Sparse Autoencoders}

Mechanistic interpretability is a research agenda which aims to understand learned model algorithms through studying their weights, (see \cite{olah2020zoom} for an introduction). Recent results relating to language models include \cite{meng2023locatingeditingfactualassociations}, which finds a correspondence between specific facts and feature weights, along with \cite{olsson2022incontextlearninginductionheads}, which shows that transformers learn in context learning through the mechanism of ``induction heads". 

\cite{elhage2022superposition} introduced the model we study in this paper. While that work focused on mapping empirically behaviors of the model in multiple regimes of interest such as correlated inputs, we focused on a regime with enough symmetry to solve the model analytically given observed symmetries in trained models. \cite{chen2023dynamicalversusbayesianphase} study this model in the context of singular learning theory. As part of their work, they characterize the loss using a different high sparsity approximation than the one we present in this paper (they assume exactly one input feature is on). Then they derive a subset of the critical points and their corresponding local learning coefficients under the assumption $n_d=2$.

While the concept of dictionary learning was introduced by \citep{mallat1993matching}, the practical use of sparse autoencoders to understand large language models has accelerated recently due to mezzo-scale open weight models \citep{gao2024scaling,lieberum2024gemmascopeopensparse} and large-scale open-output models \cite{bricken2023monosemanticity}. These features are highly interpretable \citep{cunningham2023sparse} and scale predictably. Interestingly, the scaling is quite similar for the various different architectures they consider, differing primarily by a constant, which fits with the predictions in this work. 

We have seen that the dominant source of error is not from determining which features are present, but rather the actual values of those features. Small modifications to the activation functions, such as gating \cite{rajamanoharan2024improving}, k-sparse \cite{makhzani2013k}, or TRec non-linearity \cite{ProLUNonlinearity, konda2014zero}, are insufficient to fix this problem as they do not solve the basic issue of noisy outputs. In this context our work implies that innovative architectures, that are suitable both for gradient-based training and also for decoding sparse features, must be developed.

\subsubsection{Compressed Sensing and Statistical Physics}
It is known that compressed sparse data can be exactly reconstructed by solving a convex problem \citep{ctau1, ctau2, csensingDonohoElad, donoho2006compressed} given knowledge of the compression matrix. Furthermore, using tools from statistical physics it is possible to show that this holds for typical compressed sparse data \citep{ganguli2010statistical}. Learning the compression matrix is also easy in certain circumstances\citep{sakata2013statistical}. For a more general review on compressed sensing and it's history consider the introduction by \cite{davenport2012introduction}. The reconstruction procedure typically used in compressed sensing is optimizing a (convex) relaxation of finding the sparsest set of features which reproduces your data vector. This is significantly different to the setting of sparse autoencoders which try to obtain the sparse features using only one linear + activation layer. 

The discrepancy between the ability of convex optimization techniques to achieve zero loss while a linear + ReLU model necessarily incurs an error suggests that a more complex model architecture is needed for sparse autoencoders when it is desirable to calculate the feature magnitude to high precision. This may occur, for example, if one wishes to insert a sparse autoencoder into a model without corrupting its downstream outputs.

\subsubsection{Linear Models}

The linear version of the toy model we studied has garnered significant attention. \cite{plaut2018principalsubspacesprincipalcomponents} demonstrates that the minimizers of \cref{eq::basic_loss} can be directly related to Principal Component Analysis (PCA), a fundamental technique in dimensionality reduction and data analysis (see also \cite{bishop2006pattern} for a comprehensive treatment). \cite{kunin2019loss} studies the learning dynamics of linear autoencoders when regularization is added.

\subsection{Remaining questions and future directions}

$\textbf{Correlated Data}$\\
We expect real world data to display significant correlations between activated features. Nevertheless, we expect there are more realistic correlation structures that remain simple enough to analyze the loss analytically. One such possibility is to use data whose correlations are described by a sparse graph.

$\textbf{Compressed Computation}$\\
We have shown that features can only be recovered using linear + element-wise or selection type activations together with a small error. How then are neural networks able to compute on information stored in superposition using just linear + ReLU layers? Noise injected into data early on in a computation tends to be compounded as the computation grows in circuit complexity. Thus we believe the neural networks must be at least one of the following: 
\begin{enumerate}
    \item  performing computation on superposed information without first localizing it,
    \item performing computation on superposed information by first localizing it using multiple layers,
    \item performing only low complexity computation on superposed information, or 
    \item implementing a fault tolerance scheme (computation which actively corrects errors).
\end{enumerate}

It would be interesting to experimentally look for these phenomena and their structure. 

\section*{Acknowledgements}
We are grateful for the valuable feedback we received from Daniel Kunin, David Barmherzig, Henrik Marklund, Surya Ganguli, and Zach Furman. We acknowledge Adam Brown in particular for insightful remarks that helped us construct the Persian rug matrix. Aditya Cowsik acknowledges funding from the Simons Foundation (Award ID: 560571).


\clearpage
\bibliography{iclr2025_conference}
\bibliographystyle{iclr2025_conference}
\clearpage

\appendix

\section{Optimizing over $a$}
\label{apx: optimizing-A}

We may optimize over $a$ analytically because it appears almost quadratically in the loss. Consider the expression for the average loss from \cref{eq:expected_loss} with $b$ replaced with $a \cdot b$. As long as $a \neq 0$ this redefinition doesn't change the set of accessible models.

Furtheremore let us restrict to positive $a$ which allows us to rewrite the loss as  
\begin{equation}
    L = \mathbb{E}_{x,\noise}[\left(x-\relu\left(a(x + \noise + b)\right))^2] = \mathbb{E}_{x,\noise}[(x-a \relu\left(x + \noise + b\right)\right)^2].
\end{equation}
The restriction to positive $a$ is acceptable because we never see negative off-diagonal elements in our trained models. Now, optimizing over $a$ is exactly linear regression; we can obtain the optimal value of $a$ with the standard method
\begin{align}
    0 = \frac{d}{da} L = -2 \mathbb{E}\left[ (x - a \relu (x + \noise + b)) \relu(x + \noise + b)\right]
\end{align}
which implies that 
\begin{equation}
    a_{\text{opt}} = \frac{\mathbb{E}_{x, \noise}\left[x \relu(x+\noise+b)\right]}{\mathbb{E}_{x, \noise} \left[ \relu(x + \noise + b)^2 \right]}.
\end{equation}
Notice that the optimal $a$ is always positive, which is consistent with the assumption we made earlier. 

\section{Bounding the Reconstruction Error}
\label{app:reconstruction_error}

\textbf{On term:}

We can start to write the on term similarly as
\begin{align*}
     \langle(u-\relu\left(u+\noise)\right)^2\rangle
     =
     \left\langle\sigma^2\fullint d\noise\gaussmeasrescaled (u-\relu(u+\sigma\noise))^2\right\rangle_u.
\end{align*}
Now we write the integral in two parts to get rid of the ReLU: one when $u+\sigma\noise <0$ and one when $u+\sigma\noise > 0.$ This gives
\begin{align*}
    \underbrace{
    \left\langle\int_{-\infty}^{-\frac{u}{\sigma}} d\noise\gaussmeasrescaled u^2\right\rangle_u
    }_{E_r}
    +
    \underbrace{
    \left\langle\sigma^2\int_{-\frac{u}{\sigma}}^{\infty} d\noise\gaussmeasrescaled \noise^2\right\rangle_u
    }_{E_{\noise}}.
\end{align*}

Where the first term $E_r$ represents error coming from the $\relu$ and the second term $E_\noise$ represents error coming from the noise. The scaling of $E_\noise$ can be easily bounded: 
\begin{align*}
      E_\noise  < \sigma^2\fullint d\noise\gaussmeasrescaled \noise^2 \sim O(\sigma^2 + \sigma\mu + \mu^2)
\end{align*}
And thus we see, unsurprisingly, that we need to set $\mu\ll 1$ to get a good bound. 

To upper bound  $E_r$ write the $u$ integral in two intervals: $[0,2|\mu|]$ and $[2|\mu|,1]$, corresponding to regions in which the interval of the $\noise$ integral does and "decisively" does not include the the mean respectively.  In particular, we have

\begin{align*}
     E_r <
    \underbrace{
    \int_0^{2|\mu|}u^2du  \int_{-\infty}^{-\frac{u}{\sigma}} d\noise\gaussmeasrescaled 
    }_{E_r^\text{mean}}
    +
    \underbrace{
    \int_{2|\mu|}^{1}u^2du   \int_{-\infty}^{-\frac{u}{\sigma}} d\noise\gaussmeasrescaled
    }_{E_{r}^\text{tail}}.
\end{align*}
Since in $E_r^\text{mean}$ the $\noise$ integrals' interval includes the mean we may as well extend the interval to the full real line to get the bound, giving 
\begin{align*}
    E_r^\text{mean} < \int_0^{2|\mu|}u^2du = O(\mu^3).
\end{align*}
 $E_r^\text{tail}$ can be upper bounded by setting the $\noise$ range to the maximum value of $2|\mu|$, so we have
\begin{align*}
      E_r^\text{tail} <
    (1-2|\mu|) \int_{-\infty}^{-\frac{2|\mu|}{\sigma}} d\noise\gaussmeasrescaled =  (1-2|\mu|) \int_{-\infty}^{0} d\noise \frac{e^{-\frac{(\noise-\frac{|\mu|}{\sigma})^2}{2}}}{\sqrt{2\pi}}
    <O(1)(1-2|\mu|)e^{  -\frac{|\mu|^2}{2\sigma^2}    }.
\end{align*}
Putting it all together gives
\begin{align*}
    L < 
    (1-p)O(\sigma^2 e^{-\frac{\mu^2}{2\sigma^2}}) + 
    pO(
    \mu^3 
    + 
    e^{  -\frac{|\mu|^2}{2\sigma^2}    } 
    + 
    \mu e^{  -\frac{|\mu|^2}{2\sigma^2}    } 
    + 
    \sigma^2 
    + 
    \sigma\mu 
    + 
    \mu^2).
\end{align*}
Plugging in the $\mu$ scaling from \cref{eq: mu-scaling} and keeping only the lowest order terms gives
\begin{align}
    L < O(
     \sigma^2 p\log{\frac{1}{p}} ) \sim O(\frac{p^2}{r}\log{\frac{1}{p}}).
\end{align}

\section{Minimal Variance Bound}
\label{app:min_variance}
We will show a minimum variance bound for matrices $W$ which have all diagonals equal to 1 and also have maximum rank $n_d$. In this case we know that $\Tr W = n_s$. On the other hand we also know that the trace is the sum of the eigenvalues, and because $W$ has rank at most $n_d$ that
\begin{equation}
    n_s = \sum_{i=1}^{n_d} \lambda_i
    \label{eq:diag_sum}
\end{equation}
for the eigenvalues $\lambda_i$ of $W$. Now we solve for the mean of the variance across rows,
\begin{equation}
    \frac{1}{n_s} \sum_{i=1}^{n_s} \operatorname{Var}(\nu_i) = \frac{4p-3p^2}{12n_s} \sum_{i=1}^{n_s} \sum_{j=1, j \neq i}^{n_s} W_{ij}^2 = \frac{4p-3p^2}{12n_s}\left(\Tr(WW^\dagger) - n_s\right).
\end{equation}
Here the first equality arises from the definition of $\noise_i$ (remembering that we have set the diagonals to 1 exactly) and substituting the variance of $x_j$, while the second equality follows because $\Tr(WW^\dagger)$ is the sum of the square of all entries of $W$, and we subtract off the diagonal entries. 

Because we want a bound on this quantity related to the eigenvalues of $W$, it is convenient to use the Schur decomposition of $W = QUQ^\dagger$. Here $Q$ is a unitary matrix and $U$ is upper-triangular with the eigenvalues of $W$ on the diagonal. This allows us to lower bound the trace
\begin{equation}
    \Tr(WW^\dagger) = \Tr(QUQ^\dagger Q U^\dagger Q^\dagger) = \Tr(UU^\dagger) = \sum_{i, j= 1}^{n_s} |U_{ij}|^2 \geq \sum_{i=1}^{n_d} |\lambda_i|^2 \geq \frac{n_s^2}{n_d}
\end{equation}
where the last inequality follows from Cauchy-Schwarz and \cref{eq:diag_sum}. With this we find a bound on the variance
\begin{equation}
    \frac{1}{n_s} \sum_{i=1}^{n_s} \operatorname{Var}(\noise_i) \geq \frac{4p-3p^2}{12}\left(\frac{n_s}{n_d} - 1\right),
\end{equation}
with equality if $W$ is symmetric with all non-zero eigenvalues equal. These two conditions follow because the two inequalities in the proof become equalities when these conditions are met. This naturally leads to a candidate for the optimal choice of $W$, namely matrices of the form
\begin{equation}
    W \propto O P O^T \text{ and } W_{ii} = 1
\end{equation}
where $O$ is an orthogonal matrix and $P$ is any rank-$n_d$ projection matrix. This kind of matrix saturates both bounds because it is symmetric and has all nonzero eigenvalues equal to 1.

\section{Lower bound on loss scaling}
\label{apx: lower_bounding_loss}

We now show a lower bound on the loss in the $\pf\rightarrow 0$ limit. To do this, we will show a more general lower bound on the on loss for any deterministic function of the pre-activation. Specifically, we would like to lower bound

\begin{align*}
    L \leq p L_\text{on} &=  \mathbb{E}[(u-f\left(u+\noise)\right)^2]
\end{align*}

for any function $f$ with $u\sim \text{Uniform}[0,1]$ and $\nu\sim \mathcal{N}(0,\sigma)$. Recall that there is no need to consider the bias $b$ as it can be absorbed into $\nu$. Recall that the optimal function $f$ is given by 
\begin{align*}
    f^*(u+\nu) = \mathbb{E}[u|u+\nu].
\end{align*}
Let's make a change of variable from $u, \nu$ to $y \equiv u + \nu, u$, and then use the tower property to rewrite $L_{\text{on}}$ as
\begin{equation}
    L_{\text{on}} = \mathbb{E}_{y \sim P_{u+\nu}} \left[ \mathbb{E}_{u|y}\left[\left(u - f^*(u+\nu)\right)^2\right]\right].
\end{equation}
We first draw $y$ from the marginal distribution of $u + \nu$ and then draw $u$ from the conditional distribution given $y$. Because $f^*$ is exactly the conditional expectation the interior expectation becomes the conditional variance 
\begin{equation}
    L_{\text{on}} = \mathbb{E}_{y} \left[ \operatorname{Var}\left[u | y\right]\right].
\end{equation}

Because we want to lower bound $L_{\text{on}}$ it will be convenient to start with a lower bound for the conditional variance. We will lower bound the conditional variance for $y \in [\sigma, 1-\sigma]$, and then use that lower bound to find a lower bound for the loss, with a goal of showing that the loss is lower bounded by a constant multiple of $\sigma^2$, for $\sigma < \frac{1}{4}$. This will show that the overall loss of any strategy, even one which can perfectly estimate which features are on or off, is incapable of achieving a reconstruction error better than $O(p^2 / r)$. 

The conditional distribution for $u$ is a truncated Gaussian distribution. By Bayes' theorem
\begin{align}
    P[u | u + \nu = y] &= \frac{P[u + \nu = y]P[u]}{P[y]} \\
    &= \begin{cases}
        \frac{e^{-(u - y)^2 / 2\sigma^2}}{\int_0^1 dx e^{-(x - y)^2 / 2\sigma^2}} &\text{if } u \in [0, 1] \\
        0 &\text{otherwise},
    \end{cases}
\end{align}

with normalizing constant $Z(y) = \int_0^1 dx e^{-(x - y)^2 / 2\sigma^2} < \sqrt{2\pi\sigma^2}$. This is a truncated Gaussian distribution. Fix $y \in [\sigma , 1-\sigma]$ so that all distributions are implicitly conditioned on $y$ for now. Sample $u$ via the following procedure. First we decide if $|u - y| \leq \sigma$ and then we either sample from the conditional distribution $P[u | y\text{ and } |u - y| \leq \sigma]$ or $P[u | y\text{ and } |u - y| \geq \sigma]$ with their corresponding probabilities. Let $R$ be the indicator random variable denoting $|u-y| \leq \sigma$. Then by the law of total variance 
\begin{align}
     \operatorname{Var}\left[u ~|~ y\right] &= P[R = 1] \operatorname{Var}\left[u ~|~ R=1\right] +
     P[R=0] \operatorname{Var}\left[u ~|~ R=0\right] + 
     \operatorname{Var}_R \left[ \mathbb{E}[u ~|~ R] \right] \\
     &\geq P[R = 1] \operatorname{Var}\left[u ~|~ R=1\right]
\end{align}
where we have dropped the latter two positive terms to derive the lower bound.  $P[R = 1] \geq \operatorname{erf}(2^{-1/2}) $ because the chance a truncated Gaussian is within one $\sigma$ of its mode is larger than that for an untruncated Gaussian, given that the truncation is more than $\sigma$ away from the mode. This condition is satisfied by construction because we have chosen $y$ to be more than $\sigma$ from the boundary.

Additionally a trivial scaling argument shows that the variance is proportional to $\sigma^2$ which means that there is some constant, $C > 0$ such that
\begin{equation}
    \operatorname{Var}\left[u ~|~ y\right] \geq C \sigma^2
\end{equation}
when $y \in [\sigma, 1-\sigma]$. To complete the argument we now return to 
\begin{equation}
    L_{\text{on}} = \mathbb{E}_{y} \left[ \operatorname{Var}\left[u | y\right]\right] \geq \mathbb{E}_{y} \left[ \operatorname{Var}\left[u | y\right] 1_{y \in [\sigma, 1-\sigma]} \right] \geq C \sigma^2 P[y \in [\sigma, 1-\sigma]].
\end{equation}

For $\sigma = 1/4$ this probability is clearly finite and for $\sigma < 1/4$ it is increasing as $\sigma$ decreases so it is uniformly bounded below by a constant $C_1$. So finally
\begin{equation}
    L_{\text{on}} \geq C'\sigma^2 \implies L \geq C' p\sigma^2 \approx \frac{C'p^2}{r}.
\end{equation}


\end{document}

%% file: math_commands.tex
\usepackage{amsmath,amsfonts,bm}

\def\eqref#1{equation~\ref{#1}}

\def\1{\bm{1}}

\DeclareMathAlphabet{\mathsfit}{\encodingdefault}{\sfdefault}{m}{sl}
\SetMathAlphabet{\mathsfit}{bold}{\encodingdefault}{\sfdefault}{bx}{n}

\DeclareMathOperator{\Tr}{Tr}

%% file: main.bbl
\begin{thebibliography}{28}
\providecommand{\natexlab}[1]{#1}
\providecommand{\url}[1]{\texttt{#1}}
\expandafter\ifx\csname urlstyle\endcsname\relax
  \providecommand{\doi}[1]{doi: #1}\else
  \providecommand{\doi}{doi: \begingroup \urlstyle{rm}\Url}\fi

\bibitem[Bishop \& Nasrabadi(2006)Bishop and Nasrabadi]{bishop2006pattern}
Christopher~M Bishop and Nasser~M Nasrabadi.
\newblock \emph{Pattern recognition and machine learning}, volume~4.
\newblock Springer, 2006.

\bibitem[Bricken et~al.(2023)Bricken, Templeton, Batson, Chen, Jermyn, Conerly,
  Turner, Anil, Denison, Askell, Lasenby, Wu, Kravec, Schiefer, Maxwell,
  Joseph, Hatfield-Dodds, Tamkin, Nguyen, McLean, Burke, Hume, Carter,
  Henighan, and Olah]{bricken2023monosemanticity}
Trenton Bricken, Adly Templeton, Joshua Batson, Brian Chen, Adam Jermyn, Tom
  Conerly, Nick Turner, Cem Anil, Carson Denison, Amanda Askell, Robert
  Lasenby, Yifan Wu, Shauna Kravec, Nicholas Schiefer, Tim Maxwell, Nicholas
  Joseph, Zac Hatfield-Dodds, Alex Tamkin, Karina Nguyen, Brayden McLean,
  Josiah~E Burke, Tristan Hume, Shan Carter, Tom Henighan, and Christopher
  Olah.
\newblock Towards monosemanticity: Decomposing language models with dictionary
  learning.
\newblock \emph{Transformer Circuits Thread}, 2023.
\newblock
  https://transformer-circuits.pub/2023/monosemantic-features/index.html.

\bibitem[Candes \& Tao(2005)Candes and Tao]{ctau1}
E.J. Candes and T.~Tao.
\newblock Decoding by linear programming.
\newblock \emph{IEEE Transactions on Information Theory}, 51\penalty0
  (12):\penalty0 4203--4215, 2005.
\newblock \doi{10.1109/TIT.2005.858979}.

\bibitem[Candes et~al.(2006)Candes, Romberg, and Tao]{ctau2}
E.J. Candes, J.~Romberg, and T.~Tao.
\newblock Robust uncertainty principles: exact signal reconstruction from
  highly incomplete frequency information.
\newblock \emph{IEEE Transactions on Information Theory}, 52\penalty0
  (2):\penalty0 489--509, 2006.
\newblock \doi{10.1109/TIT.2005.862083}.

\bibitem[Chen et~al.(2023)Chen, Lau, Mendel, Wei, and
  Murfet]{chen2023dynamicalversusbayesianphase}
Zhongtian Chen, Edmund Lau, Jake Mendel, Susan Wei, and Daniel Murfet.
\newblock Dynamical versus bayesian phase transitions in a toy model of
  superposition, 2023.
\newblock URL \url{https://arxiv.org/abs/2310.06301}.

\bibitem[Cunningham et~al.(2023)Cunningham, Ewart, Riggs, Huben, and
  Sharkey]{cunningham2023sparse}
Hoagy Cunningham, Aidan Ewart, Logan Riggs, Robert Huben, and Lee Sharkey.
\newblock Sparse autoencoders find highly interpretable features in language
  models.
\newblock \emph{arXiv preprint arXiv:2309.08600}, 2023.

\bibitem[Davenport et~al.(2012)Davenport, Duarte, Eldar, and
  Kutyniok]{davenport2012introduction}
Mark~A Davenport, Marco~F Duarte, Yonina~C Eldar, and Gitta Kutyniok.
\newblock Introduction to compressed sensing., 2012.

\bibitem[Donoho(2006)]{donoho2006compressed}
David~L Donoho.
\newblock Compressed sensing.
\newblock \emph{IEEE Transactions on information theory}, 52\penalty0
  (4):\penalty0 1289--1306, 2006.

\bibitem[Donoho \& Elad(2003)Donoho and Elad]{csensingDonohoElad}
David~L. Donoho and Michael Elad.
\newblock Optimally sparse representation in general (nonorthogonal)
  dictionaries via \&\#x2113;<sup>1</sup> minimization.
\newblock \emph{Proceedings of the National Academy of Sciences}, 100\penalty0
  (5):\penalty0 2197--2202, 2003.
\newblock \doi{10.1073/pnas.0437847100}.
\newblock URL \url{https://www.pnas.org/doi/abs/10.1073/pnas.0437847100}.

\bibitem[Elhage et~al.(2022)Elhage, Hume, Olsson, Schiefer, Henighan, Kravec,
  Hatfield-Dodds, Lasenby, Drain, Chen, Grosse, McCandlish, Kaplan, Amodei,
  Wattenberg, and Olah]{elhage2022superposition}
Nelson Elhage, Tristan Hume, Catherine Olsson, Nicholas Schiefer, Tom Henighan,
  Shauna Kravec, Zac Hatfield-Dodds, Robert Lasenby, Dawn Drain, Carol Chen,
  Roger Grosse, Sam McCandlish, Jared Kaplan, Dario Amodei, Martin Wattenberg,
  and Christopher Olah.
\newblock Toy models of superposition.
\newblock \emph{Transformer Circuits Thread}, 2022.
\newblock URL \url{https://transformer-circuits.pub/2022/toy_model/index.html}.

\bibitem[Engels et~al.(2024)Engels, Liao, Michaud, Gurnee, and
  Tegmark]{engels2024languagemodelfeatureslinear}
Joshua Engels, Isaac Liao, Eric~J. Michaud, Wes Gurnee, and Max Tegmark.
\newblock Not all language model features are linear, 2024.
\newblock URL \url{https://arxiv.org/abs/2405.14860}.

\bibitem[Ganguli \& Sompolinsky(2010)Ganguli and
  Sompolinsky]{ganguli2010statistical}
Surya Ganguli and Haim Sompolinsky.
\newblock Statistical mechanics of compressed sensing.
\newblock \emph{Physical review letters}, 104\penalty0 (18):\penalty0 188701,
  2010.

\bibitem[Gao et~al.(2024)Gao, la~Tour, Tillman, Goh, Troll, Radford, Sutskever,
  Leike, and Wu]{gao2024scaling}
Leo Gao, Tom~Dupr{\'e} la~Tour, Henk Tillman, Gabriel Goh, Rajan Troll, Alec
  Radford, Ilya Sutskever, Jan Leike, and Jeffrey Wu.
\newblock Scaling and evaluating sparse autoencoders.
\newblock \emph{arXiv preprint arXiv:2406.04093}, 2024.

\bibitem[Konda et~al.(2014)Konda, Memisevic, and Krueger]{konda2014zero}
Kishore Konda, Roland Memisevic, and David Krueger.
\newblock Zero-bias autoencoders and the benefits of co-adapting features.
\newblock \emph{arXiv preprint arXiv:1402.3337}, 2014.

\bibitem[Kunin et~al.(2019)Kunin, Bloom, Goeva, and Seed]{kunin2019loss}
Daniel Kunin, Jonathan Bloom, Aleksandrina Goeva, and Cotton Seed.
\newblock Loss landscapes of regularized linear autoencoders.
\newblock In \emph{International conference on machine learning}, pp.\
  3560--3569. PMLR, 2019.

\bibitem[Lieberum et~al.(2024)Lieberum, Rajamanoharan, Conmy, Smith, Sonnerat,
  Varma, Kramár, Dragan, Shah, and Nanda]{lieberum2024gemmascopeopensparse}
Tom Lieberum, Senthooran Rajamanoharan, Arthur Conmy, Lewis Smith, Nicolas
  Sonnerat, Vikrant Varma, János Kramár, Anca Dragan, Rohin Shah, and Neel
  Nanda.
\newblock Gemma scope: Open sparse autoencoders everywhere all at once on gemma
  2, 2024.
\newblock URL \url{https://arxiv.org/abs/2408.05147}.

\bibitem[Makhzani \& Frey(2013)Makhzani and Frey]{makhzani2013k}
Alireza Makhzani and Brendan Frey.
\newblock K-sparse autoencoders.
\newblock \emph{arXiv preprint arXiv:1312.5663}, 2013.

\bibitem[Mallat \& Zhang(1993)Mallat and Zhang]{mallat1993matching}
St{\'e}phane~G Mallat and Zhifeng Zhang.
\newblock Matching pursuits with time-frequency dictionaries.
\newblock \emph{IEEE Transactions on signal processing}, 41\penalty0
  (12):\penalty0 3397--3415, 1993.

\bibitem[Marks et~al.(2024)Marks, Rager, Michaud, Belinkov, Bau, and
  Mueller]{marks2024sparsefeaturecircuitsdiscovering}
Samuel Marks, Can Rager, Eric~J. Michaud, Yonatan Belinkov, David Bau, and
  Aaron Mueller.
\newblock Sparse feature circuits: Discovering and editing interpretable causal
  graphs in language models, 2024.
\newblock URL \url{https://arxiv.org/abs/2403.19647}.

\bibitem[Meng et~al.(2023)Meng, Bau, Andonian, and
  Belinkov]{meng2023locatingeditingfactualassociations}
Kevin Meng, David Bau, Alex Andonian, and Yonatan Belinkov.
\newblock Locating and editing factual associations in gpt, 2023.
\newblock URL \url{https://arxiv.org/abs/2202.05262}.

\bibitem[Olah et~al.(2020)Olah, Cammarata, Schubert, Goh, Petrov, and
  Carter]{olah2020zoom}
Chris Olah, Nick Cammarata, Ludwig Schubert, Gabriel Goh, Michael Petrov, and
  Shan Carter.
\newblock Zoom in: An introduction to circuits.
\newblock \emph{Distill}, 2020.
\newblock \doi{10.23915/distill.00024.001}.
\newblock https://distill.pub/2020/circuits/zoom-in.

\bibitem[Olsson et~al.(2022)Olsson, Elhage, Nanda, Joseph, DasSarma, Henighan,
  Mann, Askell, Bai, Chen, Conerly, Drain, Ganguli, Hatfield-Dodds, Hernandez,
  Johnston, Jones, Kernion, Lovitt, Ndousse, Amodei, Brown, Clark, Kaplan,
  McCandlish, and Olah]{olsson2022incontextlearninginductionheads}
Catherine Olsson, Nelson Elhage, Neel Nanda, Nicholas Joseph, Nova DasSarma,
  Tom Henighan, Ben Mann, Amanda Askell, Yuntao Bai, Anna Chen, Tom Conerly,
  Dawn Drain, Deep Ganguli, Zac Hatfield-Dodds, Danny Hernandez, Scott
  Johnston, Andy Jones, Jackson Kernion, Liane Lovitt, Kamal Ndousse, Dario
  Amodei, Tom Brown, Jack Clark, Jared Kaplan, Sam McCandlish, and Chris Olah.
\newblock In-context learning and induction heads, 2022.
\newblock URL \url{https://arxiv.org/abs/2209.11895}.

\bibitem[Plaut(2018)]{plaut2018principalsubspacesprincipalcomponents}
Elad Plaut.
\newblock From principal subspaces to principal components with linear
  autoencoders, 2018.
\newblock URL \url{https://arxiv.org/abs/1804.10253}.

\bibitem[Rajamanoharan et~al.(2024)Rajamanoharan, Conmy, Smith, Lieberum,
  Varma, Kram{\'a}r, Shah, and Nanda]{rajamanoharan2024improving}
Senthooran Rajamanoharan, Arthur Conmy, Lewis Smith, Tom Lieberum, Vikrant
  Varma, J{\'a}nos Kram{\'a}r, Rohin Shah, and Neel Nanda.
\newblock Improving dictionary learning with gated sparse autoencoders.
\newblock \emph{arXiv preprint arXiv:2404.16014}, 2024.

\bibitem[Sakata \& Kabashima(2013)Sakata and Kabashima]{sakata2013statistical}
Ayaka Sakata and Yoshiyuki Kabashima.
\newblock Statistical mechanics of dictionary learning.
\newblock \emph{Europhysics Letters}, 103\penalty0 (2):\penalty0 28008, 2013.

\bibitem[Taggart(2024)]{ProLUNonlinearity}
Glen~M. Taggart.
\newblock Prolu: A nonlinearity for sparse autoencoders.
\newblock
  \url{https://www.alignmentforum.org/posts/HEpufTdakGTTKgoYF/prolu-a-nonlinearity-for-sparse-autoencoders},
  2024.

\bibitem[Waldron(2018)]{waldron2018introduction}
Shayne~FD Waldron.
\newblock \emph{An introduction to finite tight frames}.
\newblock Springer, 2018.

\bibitem[Welch(1974)]{welch1974lower}
Lloyd Welch.
\newblock Lower bounds on the maximum cross correlation of signals (corresp.).
\newblock \emph{IEEE Transactions on Information theory}, 20\penalty0
  (3):\penalty0 397--399, 1974.

\end{thebibliography}
